\documentclass[letterpaper, 10 pt, conference]{ieeeconf}  

\IEEEoverridecommandlockouts                              

\usepackage{array} 
\usepackage{graphicx}
\usepackage{amsmath} 
\usepackage{comment}
\usepackage{booktabs}       
\usepackage{multirow}
\usepackage{makecell}
\usepackage{adjustbox}
\usepackage{xcolor}
\usepackage{cite}

\usepackage{bbold}
\usepackage{amsfonts}
\usepackage{amssymb}
\usepackage{txfonts}

\usepackage{booktabs}
\usepackage{nicematrix}

\newcommand{\gap}[1]{\scriptsize($+$#1)}
\newcommand{\mgap}[1]{\scriptsize($-$#1)}
\newcommand{\gapbold}[1]{\textbf{\scriptsize($+$#1)}}
\newcommand{\base}[1]{\scriptsize(0)#1}
\newcolumntype{C}{>{\centering\arraybackslash}p{4.5em}}

\title{\LARGE \bf
GVCCI: Lifelong Learning of Visual Grounding for \\Language-Guided Robotic Manipulation
}

\author{Junghyun Kim$^1$ Gi-Cheon Kang$^{1*}$ Jaein Kim$^{2*}$ Suyeon Shin$^{1}$ Byoung-Tak Zhang$^{12}$
\thanks{$^*$Authors have equal contributions}
\thanks{$^{1}$Interdisciplinary Program in AI, Seoul National University}
\thanks{$^{2}$Interdisciplinary Program in Neuroscience, Seoul National University}
\thanks{All authors are also affiliated with AI Institute, Seoul National University}
\thanks{
This research was supported by the Institute of Information \& Communications Technology Planning \& Evaluation (IITP) (2021-0-02068-AIHub/10\%, 2021-0-01343-GSAI/30\%, 2022-0-00951-LBA/20\%, 2022-0-00953-PICA/40\%) grant funded by the Korean government.
}
}

\begin{document}

\maketitle
\thispagestyle{empty}
\pagestyle{empty}

\begin{abstract}
Language-Guided Robotic Manipulation (LGRM) is a challenging task as it requires a robot to understand human instructions to manipulate everyday objects. 
Recent approaches in LGRM rely on pre-trained Visual Grounding (VG) models to detect objects without adapting to manipulation environments. 
This results in a performance drop due to a substantial domain gap between the pre-training and real-world data.
A straightforward solution is to collect additional training data, but the cost of human-annotation is extortionate.
In this paper, we propose Grounding Vision to Ceaselessly Created Instructions (GVCCI), a lifelong learning framework for LGRM, which continuously learns VG without human supervision.
GVCCI iteratively generates synthetic instruction via object detection and trains the VG model with the generated data.
We validate our framework in offline and online settings across diverse environments on different VG models.
Experimental results show that accumulating synthetic data from GVCCI leads to a steady improvement in VG by up to 56.7\% and improves resultant LGRM by up to 29.4\%.
Furthermore, the qualitative analysis shows that the unadapted VG model often fails to find correct objects due to a strong bias learned from the pre-training data. 
Finally, we introduce a novel VG dataset for LGRM, consisting of nearly 252k triplets of image-object-instruction from diverse manipulation environments.
\end{abstract}

\section{INTRODUCTION}

Integrating natural language understanding and robotic manipulation is a longstanding goal 
in the fields of 
robotics and artificial intelligence (AI). 
Developing intelligent robots that can comprehend and execute natural language instructions has numerous practical applications, such as assisting users with household chores, bringing objects, and providing care for elderly or disabled people. 
In this context, Language-Guided Robotic Manipulation (LGRM) has been studied to develop robots that can manipulate everyday objects based on human language 
instructions.

One of the critical components of LGRM is to pinpoint visual objects that human instruction describes, commonly called \textit{Visual Grounding} (VG)~\cite{refcoco,referitgame}.
Since real-world environments often contain numerous objects with similar characteristics, LGRM focuses on instructions that include diverse visual information such as object categories, attributes, and spatial relationships.
As an illustration, robots should execute instructions like ``please pick up the horse-shaped toy and place it in the green box on the shelf'' or ``could you bring me the white mug next to the vase?''.
Thus the process of VG in LGRM requires a deep understanding of diverse information in both visual perception and the semantics of natural language expressions. 
Many researchers in computer vision and natural language processing communities have made remarkable progress~\cite{ofa,mdetr,vilbert} with diverse benchmark datasets~\cite{refcoco,refcocog}.

\begin{figure}[t]
\label{fig:first_page}
\centering
\includegraphics[width=\linewidth]{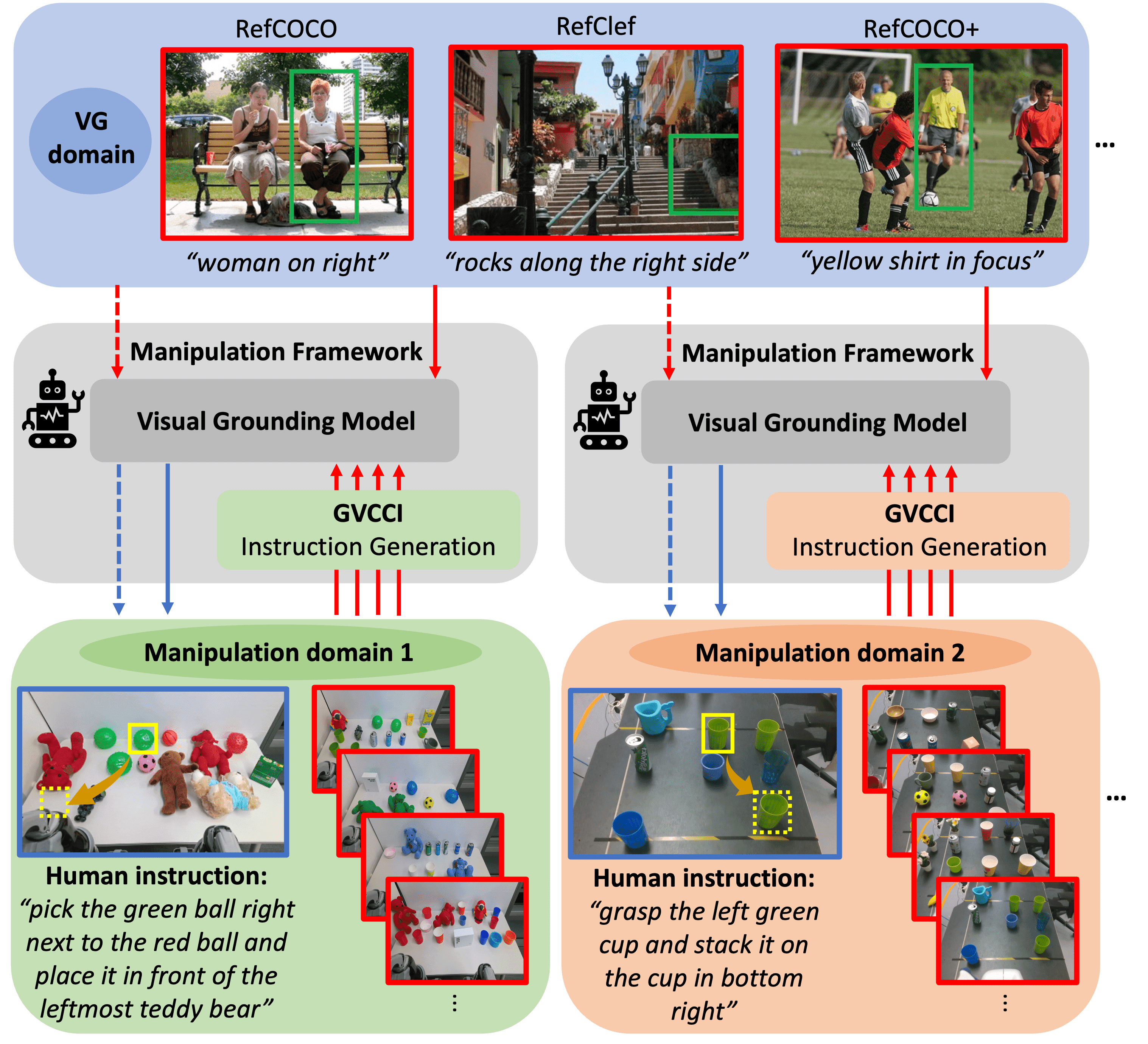}
\caption{Examples of Language Guided Robotic Manipulation (LGRM) are presented in the blue boxes along with the human instructions below. Previous work (dotted lines) has trained the VG model from the VG domain and applied it directly to the real-world manipulation domain to handle LGRM. GVCCI (solid lines), on the other hand, trains the VG model with self-generated instructions regarding observed scenes from the manipulation domain. Red lines and boxes in the figure are associated with training, and blue is associated with inference.}
\label{fig:scenario}
\end{figure}

Studies on LGRM~\cite{srfnlifrm, invigorate, ingress, lopfribhsr, cpaptbgl, avgrefrm, ivgorefhri, gssrefhri, reiofitsf} utilize deep learning-based VG models trained on large-scale datasets~\cite{refcoco,plummer2015flickr30k,referitgame}. 
However, they do not adapt the models to the real-world robotic environment as seen in the dotted lines in Fig.~\ref{fig:first_page}, assuming that the datasets for VG are a good proxy for the real world. 
We argue that applying the VG model to the real world without adaptation severely limits the manipulation ability due to the domain gap between the training datasets and data observed in the real-world environment: the properties and quantity of objects and the environment itself affect the inconsistency of the domain.
Consequently, the VG model remains poorly fitted to the robotic environment.
One immediate solution for adapting the VG model is to retrain it using VG data collected through extensive human annotation. 
However, as demonstrated by the work~\cite{iprwowusli}, manually constructing an environment-specific dataset is prohibitively expensive and arduous. Above all, a new dataset should be collected whenever the environment changes. 


To this end, we propose Grounding Vision to Ceaselessly Created Instructions (GVCCI), a lifelong learning framework for LGRM where a robot continuously learns visual grounding without human supervision. 
The core idea of our approach is automatically producing natural language instructions for manipulation via a pseudo-instruction generation method. 
Our framework first extracts the objects and their corresponding location, category, and attributes in the visual input with the off-the-shelf object detectors~\cite{fasterRCNN,bottomup} and extracts relationships between the detected objects with the proposed heuristic algorithm.
Then pseudo-instruction is generated with object features via predefined templates.
A self-generated triplet consisting of the given image, the target object's coordinates, and instructions is stored in the buffer that stochastically forgets earlier collected data.
Finally, GVCCI updates the VG model using the stored data, enabling the model to adapt to the real-world environment without requiring manual annotations.

\begin{figure}[t]
\centering
\includegraphics[width=\linewidth]{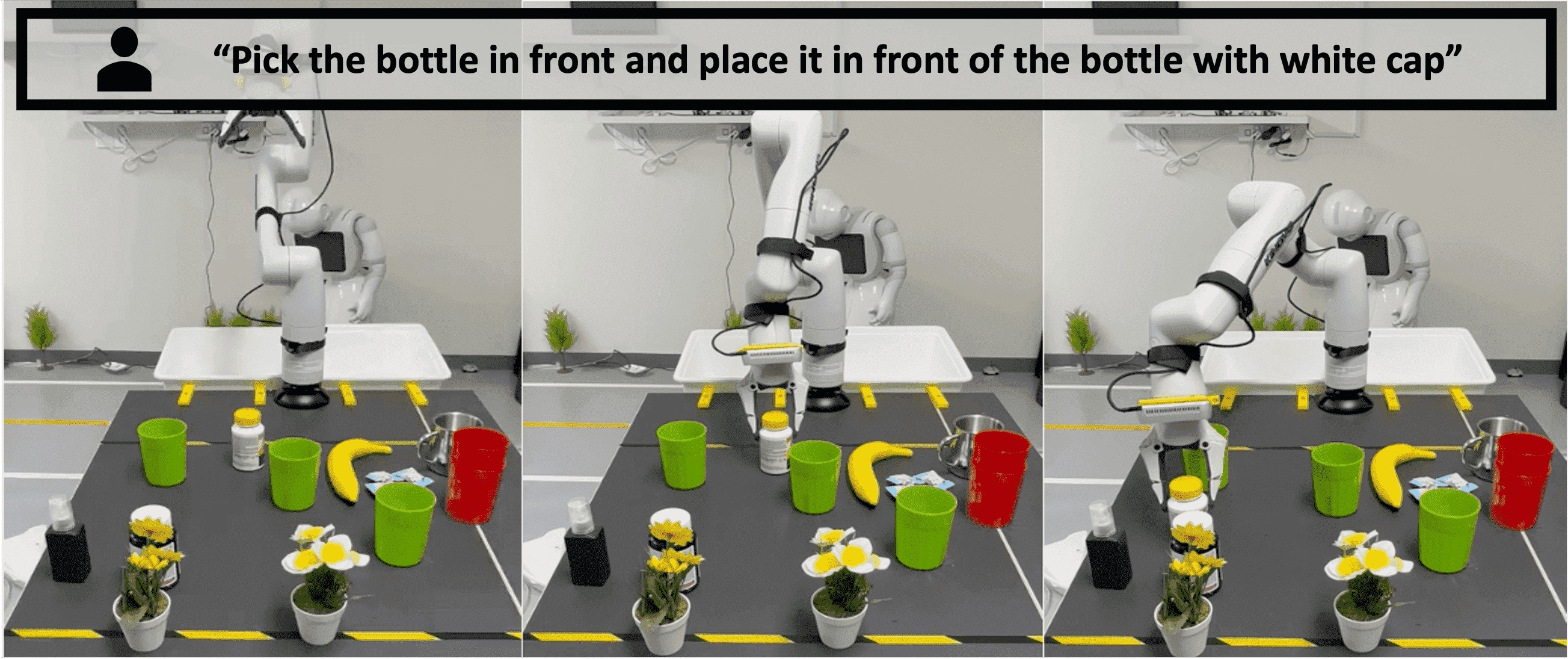}
\caption{\textbf{Language Guided Robotic Manipulation (LGRM).} We define a task of LGRM as performing \emph{pick-and-place} based on human instruction. First, the robot has to pick the referred object and place it at the target destination described in the human instruction.}
\label{fig:task}
\end{figure}

To validate the robustness and effectiveness of our approach, we collected 150 images containing numerous objects and 528 human-annotated instructions from two different environments, resulting in three distinct test sets with varying properties.
We conduct offline (VG) and online (LGRM) experiments on these datasets. 
In the offline experiment, we identify the efficacy of lifelong learning on two state-of-the-art VG models.
Regardless of the specific VG model or environment, GVCCI monotonically enhances the performance of state-of-the-art VG models through lifelong learning, with up to 56.7\% compared to the Zero-Shot (i.e., not adapted) VG model as synthetic VG data accumulates.
Furthermore, we conduct the online experiment using a real-world arm robot as depicted in Fig.~\ref{fig:task}. 
Results from the experiment show that our framework increases the manipulation performance by up to 29.4\%.

To summarize, our contributions are mainly three-fold:      

\begin{itemize}
\item We introduce GVCCI, a lifelong learning framework that enables intelligent robots to continuously learn visual grounding by generating pseudo-instructions, thereby circumventing human annotation.
\item We demonstrate the efficacy of GVCCI through experiments conducted in both online and diverse offline settings and highlight the importance of real-world adaptation for a state-of-the-art visual grounding model.
\item We propose a novel visual grounding dataset for a \textit{pick-and-place} task, $VGPI$, comprising 825 images collected from two robotic environments, including 528 human instructions and 252,420 self-generated instructions.
\end{itemize}

\section{RELATED WORK}

\subsection{Visual Grounding}

Visual grounding (VG) aims to localize objects described in the referring expression.
Various deep learning-based approaches~\cite{mattnet, ofa, mdetr, vilbert} have been proposed to tackle VG, but vision-language pre-training (VLP)~\cite{ofa,mdetr,vilbert} has emerged as the most dominant approach.
Based on the transformer architecture \cite{transformer}, VLP models are first pre-trained on a large-scale vision-language dataset to learn cross-modal semantics and then fine-tuned on a VG dataset. 
While these approaches utilize predetermined and static datasets, GVCCI stands apart by performing lifelong learning through generating synthetic expressions from continually perceived images. 

To generate synthetic expressions for VG, our method uses various object information, such as category, attribute, location, and spatial relationships. 
Especially, referring expressions for robotic manipulation often need spatial relationships (e.g., in front of, next to) to disambiguate the objects with a similar appearance. 
Thus it focuses more on generating expressions that include relational information than existing query generation methods~\cite{pseudoq, ocidref}.

\begin{figure*}[t]
\centering
\includegraphics[width=\linewidth]{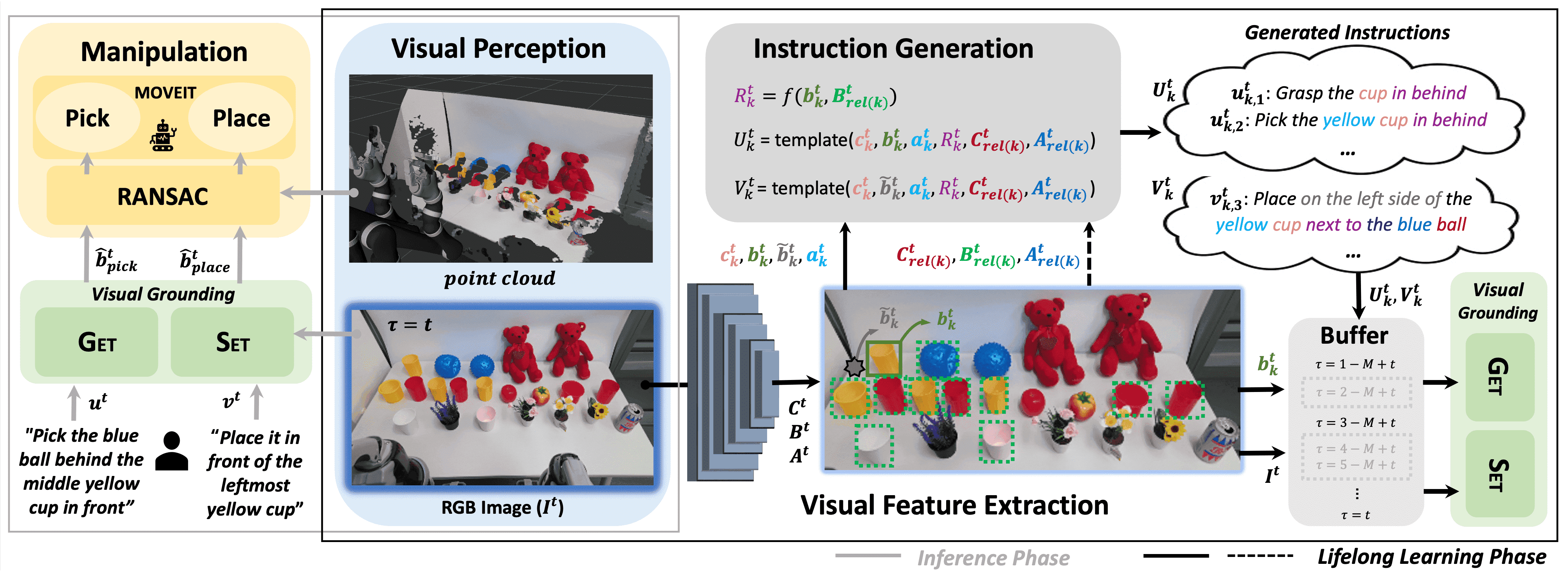}
\caption{\textbf{Overview of our lifelong learning framework.} Black solid and dotted lines are for the lifelong learning phase and gray for the inference phase. The figure illustrates an example at time step t, which can go either one of the two phases. The object with the black solid bounding box is an example of a reference object $o^t_{ref}$ that is chosen to generate the pick instructions, and the blue star on the left side of the object is an example destination $\tilde{b}^t_{ref}$.}
\label{fig:framework}
\end{figure*}

\subsection{Language-guided Robotic Manipulation}

Extensive works of Language-Guided Robotic Manipulation (LGRM) have been studied in desire of the natural interaction with robots.
We categorize these works into two streams: grounding language to action \cite{saycan,aabfigifet}
and grounding language to vision \cite{srfnlifrm, invigorate, ingress, lopfribhsr, cpaptbgl, avgrefrm, iprwowusli, ivgorefhri, gssrefhri, reiofitsf}.
We focus on the latter stream of works that aims to ground language to objects in an image to successfully manipulate referred objects. 

Although purposeful approaches have grounded language to vision in robotics, earlier works studied strong constraints. 
A work~\cite{gsrfhri} used pre-defined object categories which limit the supported instruction. 
Some studies classified manipulation targets from instructions but assumed to know the information of the objects, such as location, beforehand \cite{nlcwr, cwrumrn}. 
Thus many studies have aimed to guide a real-world manipulation task with unconstrained language instruction and unknown object information, taking different perspectives to amortize such complexity. 
A few studies \cite{gssrefhri, ivgorefhri, srfnlifrm} tried to learn spatial information in a given scene and human instructions, and \cite{reiofitsf, ivgorefhri, iprwowusli, cpaptbgl, ingress, invigorate, iogusg} tried to handle ambiguous human instructions with human-robot interaction. Other methods, such as \cite{reiofitsf, avgrefrm}, took advantage of additional non-linguistic modalities such as gesture and audio.
These works tackled an essential problem to conduct LGRM, but most of the works took advantage of a pre-trained model on existing VG datasets without any adaptation to the environment, performing manipulation tasks with zero-shot predictions of VG (See Fig.~\ref{fig:scenario}). 
This discrepancy between the training and the inference domain is mainly responsible for frequent failures of VG and subsequent failures in LGRM in the complex real world. 

A work~\cite{iprwowusli} relieved this problem by adapting the VG model with the labeled dataset of their environment. 
Nevertheless, labeling VG data for every shifted domain the robot encounters is impractical. 
A work~\cite{sim2real} has the most analogous motivation as ours, raising a problem of domain discrepancy of VG when conducting LGRM. 
They alleviated this problem by learning VG with automatically sampled referring expressions in simulation and transferring the knowledge to the real world. 
However, object categories need to be defined in simulation, and additional human labor is needed when transferring the knowledge to the real world, making robots hard to adapt in a lifelong manner.


\section{METHOD}
\label{sec:method}

GVCCI consists of 5 modules: (1) visual feature extraction module, (2) instruction generation module, (3) probabilistic volatile buffer, (4) visual grounding model, and (5) manipulation module. The overall framework is shown in Fig.~\ref{fig:framework}. 

\noindent \textbf{Notations.}
The robot's visual perception at time step $t$ consists of $I^t$, the 2D RGB input, and the corresponding 3D $point$ $cloud$.
The visual perception contains $K$ objects $O^t = \{o_k^t\}_{k=1}^K$. 
We also define $O_{rel(k)}^t\subset O^t$ as a set of objects that are associated with $o_{k}^t$. Objects in $O_{rel(k)}^t$ is either an object nearby $o_{k}^t$ or has the same object category as $o_{k}^t$. Examples of objects in $O_{rel(k)}^t$ are shown with the dotted bounding boxes in Fig.~\ref{fig:framework} where $o_{k}^t$ is depicted with the solid bounding box. 
Our approach generates the instructions for picking $U_{k}^t=\{u_{k,l}^t\}_{l=1}^L$ and placing $V_{k}^t=\{v_{k,l}^t\}_{l=1}^L$ regarding $o_k^t$. In the inference phase, the robot grounds the location of the target object $\hat{b}_{pick}^t$ and target place $\hat{b}_{place}^t$ utilizing visual information $I^t$ and human instructions $u^t$, $v^t$.

\subsection{Visual Feature Extraction Module}

The visual feature extraction module takes $I^t$ from the visual perception and extracts three visual features for each object in $O^t$: category features $C^t$, location features $B^t$, and attribute features $A^t$. 
First, Faster RCNN~\cite{fasterRCNN} is used to extract the category features $C^t=\{c_k^t\}_{k=1}^K$ and the location features $B^t=\{b_k^t\}_{k=1}^K$. Each element in the location features (i.e., $b^t_k$) consists of the top-left and bottom-right coordinates of the object bounding box. 
Meanwhile, bottom-up-attention \cite{bottomup} extracts object attribute features $A^t=\{a_k^t\}_{k=1}^K$. We follow the type of category and attribute features defined in the Visual Genome dataset \cite{visual_genome}, consisting of 1600 object categories and 400 attributes (e.g., color and material). 

\subsection{Instruction Generation Module}
\label{ssec:igm}
Inspired by the success of pseudo-supervision generation methods \cite{pseudoq, ocidref}, our module aims to generate plausible pick-and-place instructions with the features obtained from the visual feature extraction module.
When generating referring expressions for $o_{k}^t$, visual features regarding $o_{k}^t$ and $O_{rel(k)}^t$ are used, denoted as $c_{k}^t,b_{k}^t,a_{k}^t$ and $C_{rel(k)}^t,B_{rel(k)}^t,A_{rel(k)}^t$, respectively. 
Additionally, using the location features $b_{k}^t$ and $B_{rel(k)}^t$, a set of spatial relational features $R_{k}^t$ with related objects are extracted with the proposed heuristic algorithm, $f$. 
The algorithm compares $b_{k}^t$ and $B_{rel(k)}^t$ and draws three types of relational features $R_{k}^t$: (1) relationship with objects of the same category (e.g., $u^t_{k,1}$: ``cup \textit{in behind}" in Fig.~\ref{fig:framework}), (2) relationship with objects of the same category and attribute (e.g., $u^t_{k,2}$: ``yellow cup \textit{in behind}" in Fig.~\ref{fig:framework}), and (3) relationship with objects nearby the referred object (e.g., $v^t_{k,3}$: ``yellow cup \textit{next to} the blue ball" in Fig.~\ref{fig:framework}). 
Using the features $c_{k}^t,b_{k}^t,a_{k}^t,R_{k}^t,C_{rel(k)}^t,A_{rel(k)}^t$, referring expressions for each object are generated through the pre-defined templates in Tab.~\ref{tab:template}.
The full instruction is built with command terms and expressions. For example, pick instruction $u^t_{k,2} \in U_k^t$ in Fig.~\ref{fig:framework} is generated by $<$ $\text{``Pick''}+\text{Template \uppercase\expandafter{\romannumeral3}}$ $>$. 
For generating place instructions $V_{k}^t$, prepositions (e.g., in front of) are also attached to the expression. For example, $v^t_{k,3} \in V^t_k$ in Fig.~\ref{fig:framework} is generated by $<$ $\text{``Place''}+\text{\{preposition\}}+\text{Template \uppercase\expandafter{\romannumeral4}}$ $>$\footnote{Note that the place instruction generated here is not for placing the object $o_k^t$ but to place the target object in the pick instruction, $o_{i}^t \in \left(O^t \setminus \{ o_{k}^t \}\right)$, to the destination nearby $o_k^t$. See the example instruction $v^t_{k,3}$ in Fig.~\ref{fig:framework}.}. 
Our approach produces the synthetic instructions for all $K$ objects in the given image, constructing $U^t={\{U_{k}^t\}_{k=1}^K}$ and $V^t={\{V_{k}^t\}_{k=1}^K}$.

\begin{table}[t]
\caption{Referring Expression Template}
\label{tab:template}
\begin{center}
\begin{tabular}{ccc}
\toprule
 & \textbf{Template} & \textbf{Example} 
\\
\midrule
\uppercase\expandafter{\romannumeral1} & \{a\}+\{c\}                   & “wooden bowl”                 
\\
\midrule
\uppercase\expandafter{\romannumeral2} & \begin{tabular}[c]{@{}c@{}}\{R\}+\{c\}\\ \{c\}+\{R\}\end{tabular} &
  \begin{tabular}[c]{@{}c@{}}“rightmost can”\\ “can in front”\end{tabular} 
\\
\midrule
\uppercase\expandafter{\romannumeral3} & \begin{tabular}[c]{@{}c@{}}\{R\}+\{a\}+\{c\}\\ \{a\}+\{c\}+\{R\}\end{tabular} &
  \begin{tabular}[c]{@{}c@{}}“left yellow can”\\ “yellow can in behind”\end{tabular} 
\\
\midrule
\uppercase\expandafter{\romannumeral4} & \begin{tabular}[c]{@{}c@{}}\{c\}+\{R\}+\{A\}+\{C\}\\ \{a\}+\{c\}+\{R\}+\{A\}+\{C\}\end{tabular} &
  \begin{tabular}[c]{@{}c@{}}“can next to blue box”\\ “yellow can on the right of blue box”\end{tabular}
\\
\midrule
\uppercase\expandafter{\romannumeral5} & \begin{tabular}[c]{@{}c@{}}\{c\}+\{b\}\\ \{a\}+\{c\}+\{b\}\end{tabular} &
  \begin{tabular}[c]{@{}c@{}}“can on the top left of the table”\\ “yellow can on the center”\end{tabular}\\
\bottomrule
\end{tabular}
\end{center}
\end{table}

\subsection{Probabilistic Volatile Buffer}
An image $I^t$, collected instructions $U^t$ and $V^t$, and bounding box features $B^t$ are saved to the buffer later to be used to train the VG model. All collected data at $\tau=0,1,...,t$ should remain and be used for training. 
However, since our method assumes learning in a lifelong manner ($t \to \infty$) with a finite memory size of the hardware, our buffer has a mechanism that forgets the formerly generated data when the buffer overflows with the data. 
We implement the buffer to forget the earlier data with the exponential forgetting probability as follows: 
\begin{equation}
\label{eq:buffer}
\begin{aligned}
p_{forget}(\tau, t)=\min(\frac{e^{-\gamma(\tau-t)}-1}{e^{\gamma M}-1},1),
\end{aligned}
\end{equation}
where $M$ indicates the maximum number of images the buffer can hold, $t$ is the current time step. 
At time step t, the probability of the data before time step $\tau=t-M$ gives 1, which implies that the data before this time step is deleted.
From data collected after the time step $\tau=t-M$, the probability exponentially decreases as 
$\tau$ increase and reaches to 0 when $\tau=t$.
$\gamma \geq 0$ is a hyperparameter that controls how fast the forgetting probability should decay.
To summarize, the buffer will try to forget the earlier data collected,
maintaining the maximum size of the buffer to $M$.

\begin{table*}[t]
\caption{
\begin{flushleft}
\textnormal{\textbf{Results of offline experiments.}
Results show the Zero-Shot (Z-S, time step 0) performance of backbone models (Back.), and the performance of backbone models updated with two different approaches (PseudoQ and GVCCI) at selected time steps (8, 33, 135, 540).
Bold numbers indicate the best performance among time steps achieved by each approach for a given backbone model.
Numbers in parentheses represent the absolute improvement in score compared to the backbone models' Zero-Shot performance.
}
\end{flushleft}
}

\label{tab:offline_result}
\begin{center}
\adjustbox{width=\textwidth}{
\begin{NiceTabular}{|c|c|c|cccc|cccc|}
\toprule
 \multirow{2}{*}{\textbf{\rotatebox[origin=c]{90}{Back.}}}&\multirow{2}{*}{\textbf{Dataset}}& \textbf{Z-S} & \multicolumn{4}{c}{\textbf{PseudoQ~\cite{pseudoq}}} & \multicolumn{4}{c}{\textbf{GVCCI (Ours)}}
\\ 

\cmidrule(lr){3-3} \cmidrule(lr){4-7} \cmidrule(lr){8-11}
 & & \textbf{0} &\textbf{8} &\textbf{33} &\textbf{135} &\textbf{540} & \textbf{8} &\textbf{33} &\textbf{135} &\textbf{540} 
\\ 
\midrule
\textbf{\rotatebox[origin=c]{90}{MDETR}}

& \begin{tabular}[c]{@{}c@{}} Test-H \\ Test-R \\ Test-E\end{tabular}
& \begin{tabular}[c]{@{}c@{}}14.6  \\ 8.3  \\ 47.1  \end{tabular}
& \begin{tabular}[c]{@{}c@{}}58.0 \gap{43.4} \\  56.7 \gap{48.4} \\ 54.4 \gap{7.3} \end{tabular}
& \begin{tabular}[c]{@{}c@{}} 60.4 \gap{45.8} \\ \textbf{60.0} \gapbold{51.7} \\ \textbf{58.8} \gapbold{11.7} \end{tabular}
& \begin{tabular}[c]{@{}c@{}}59.4 \gap{44.8}\\ 53.9 \gap{45.6} \\ 57.4 \gap{10.3} \end{tabular} 
& \begin{tabular}[c]{@{}c@{}} \textbf{61.3} \gapbold{46.7}\\ 58.9 \gap{50.6} \\ - \end{tabular}

& \begin{tabular}[c]{@{}c@{}} 49.5 \gap{34.9}\\50.0 \gap{41.7} \\ 66.2 \gap{19.1}\end{tabular}
& \begin{tabular}[c]{@{}c@{}} 57.6 \gap{43.0}\\57.2 \gap{48.9} \\ 69.1 \gap{22.0}\end{tabular}
& \begin{tabular}[c]{@{}c@{}} 65.6 \gap{51.0}\\64.4 \gap{56.1} \\ \textbf{72.1}\gapbold{25.0}\end{tabular}
& \begin{tabular}[c]{@{}c@{}} \textbf{66.5} \gapbold{51.9}\\\textbf{65.0} \gapbold{56.7} \\ - \end{tabular}
\\ 
\midrule
\textbf{\rotatebox[origin=c]{90}{OFA}} 
&\begin{tabular}[c]{@{}c@{}} Test-H \\ Test-R \\ Test-E\end{tabular}
& \begin{tabular}[c]{@{}c@{}} 26.1  \\ 32.4  \\ 55.2  \end{tabular}
& \begin{tabular}[c]{@{}c@{}} 45.0 \gap{18.9}\\54.8 \gap{22.4} \\50.8 \mgap{4.4}\end{tabular}
& \begin{tabular}[c]{@{}c@{}} 55.5 \gap{29.4}\\61.5 \gap{29.1} \\46.3 \mgap{8.9}\end{tabular}
& \begin{tabular}[c]{@{}c@{}} \textbf{68.7} \gapbold{42.6}\\ \textbf{65.4} \gapbold{33.0} \\ \textbf{65.7} \gapbold{10.5}\end{tabular}
& \begin{tabular}[c]{@{}c@{}} 66.8 \gap{40.7}\\60.9 \gap{28.5} \\ - \end{tabular}
& \begin{tabular}[c]{@{}c@{}} 54.5 \gap{28.4}\\ 61.5 \gap{29.1} \\ 53.7 \mgap{1.5}\end{tabular}
& \begin{tabular}[c]{@{}c@{}} 69.7 \gap{43.6}\\71.5 \gap{39.1}\ \\67.7 \gap{12.5}\end{tabular}
& \begin{tabular}[c]{@{}c@{}} 72.0 \gap{45.9}\\75.4 \gap{43.0} \\ \textbf{79.1} \gapbold{23.9} \end{tabular}
& \begin{tabular}[c]{@{}c@{}} \textbf{74.4} \gapbold{48.3}\\\textbf{77.7} \gapbold{45.3} \\ -\end{tabular}
\\ 

\bottomrule
\end{NiceTabular}
}
\end{center}
\end{table*}

\subsection{Visual Grounding Model}
\label{ssec:vgm}
The Visual Grounding (VG) module utilizes pick-and-place instructions to infer the target object to pick and the destination to place. Thus, we train two separate models, Grounding Entity Transformer (\textsc{Get}) and Setting Entity Transformer (\textsc{Set}). \textsc{Get} is a Transformer \cite{transformer} based model that grounds the object entity to pick, and \textsc{Set} is another Transformer based model that infers the destination to place the object. 

As in Sec.~\ref{ssec:igm}, the instruction generation module produces the place instructions using prepositions. Accordingly, the bounding box coordinates $B^t$ should be shifted considering the preposition. For example, for the place instruction ``place it \emph{in front of} the red cup", the corresponding bounding box still refers to the red cup. We thus shift the bounding box to the front of the red cup. Likewise, all bounding boxes $B^t$ for place instructions are turned into $\tilde{B}^t$.
Based on the backbone models~\cite{ofa, mdetr}, we separately train \textsc{Get} and \textsc{Set} with $(I^t, U^t,{B}^t)$ and $(I^t, V^t,\tilde{B}^t)$, respectively. Note that the remaining data in the probabilistic volatile buffer is used for training. Finally, we optimize \textsc{Get} and \textsc{Set} by minimizing the negative log-likelihood of the bounding box as follows:
\begin{equation}
\label{eq:get_loss}
\begin{aligned}
\mathcal{L}_{\textsc{Get}}=-\sum_{d \in D}^{}\sum_{k=1}^{K}\sum_{l=1}^{L}logP_{\theta_{1}}(b^d_k|I^d,u^d_{k,l});
\end{aligned}
\end{equation}

\begin{equation}
\label{eq:set_loss}
\begin{aligned}
\mathcal{L}_{\textsc{Set}}=-\sum_{d \in D}^{}\sum_{k=1}^{K}\sum_{l=1}^{L}logP_{\theta_{2}}(\tilde{b}^d_k|I^d,v^d_{k,l}).
\end{aligned}
\end{equation}


At the inference phase, \textsc{Get} and \textsc{Set} infer the location of the referred object  $\hat{b}^t_{pick}$ and the destination $\hat{b}^t_{place}$, given $I^t$ and pick and place instructions $(u^t,v^t)$.

\subsection{Manipulation Module}

The manipulation module takes predicted 2D bounding boxes $\hat{b}^t_\text{pick}$ and $\hat{b}^t_{place}$ from the visual grounding model and computes the 3D pick-and-place coordinates. 
It first matches the 2D bounding boxes to the \textit{point cloud} with the identical resolution as $I^{t}$ and segments points of the object inside the boxes using RANSAC\cite{wahl2005identifying}.
Computing the target coordinates by averaging the estimated coordinates of segmented points,
the module plans the actual manipulation trajectory of the robotic arm using the MoveIt\cite{coleman2014moveit} software.


\section{EXPERIMENTS}
Our method is evaluated in offline and online settings and compared to various baseline methods. 
In offline settings, we conduct a comprehensive analysis to demonstrate the proficiency of the VG model (i.e., \textsc{Get}). 
In online settings, we validate GVCCI on language-guided robotic manipulation (LGRM) in \textsc{Env2} with a real-world arm robot.

\subsection{Experiment Setup}
\label{ssec:dataset}

\noindent\textbf{Environment setup.} We built two environments, \textsc{Env1} and \textsc{Env2}, scattering arbitrary everyday objects on the table. 
These environments were designed to imitate real-world scenarios, such as a table, kitchen, or room, intentionally populating with objects of similar categories and attributes. 
In most cases in such environments, a simple referring expression is insufficient for the unique identification of an object.
For example, in Fig.~\ref{fig:task}, ``green cup'' is not enough to identify the cup to which the expression refers. Two environments are different in many ways, such as objects, background, robotic hardware, and robots' viewpoint, including the angle and distance from the objects. 
Particularly in \textsc{Env2}, the portion of the object area is about three times larger than those in \textsc{Env1} on average (see Sec.~\ref{ssec:qualitative}).

\noindent\textbf{Dataset.} We construct a novel VG dataset $VGPI$ (\textit{Visual Grounding on Pick-and-place Instruction}), which contains images from both \textsc{Env1} and \textsc{Env2}. 
First, \textsc{Env1} data consists of two test sets (i.e., Test-H and Test-R) and a train set. 
Test-H includes 60 images and corresponding 212 instructions and the target object's bounding box coordinates, and Test-R has 60 images with 180 instructions and coordinates. 
The images in Test-H contain a relatively \textbf{H}igher number of objects, with an average of 11.4 objects, while the images in Test-R contain a \textbf{R}educed number of objects, with an average of 7.4.
We construct the train set with 540 images and the generated instructions, 97,448 for \textsc{Get} and 114,422 for \textsc{Set}. 
\textsc{Env}2 data consists of Test-E (E stands for \textbf{E}nlarged object size in \textsc{Env}2) and a train set. 
Test-E includes 30 images containing 7.5 objects on average, along with 68 instructions and the bounding box coordinates. The train set for \textsc{Env2} has 135 images and corresponding synthetic instructions; 19,511 for \textsc{Get} and 20,999 for \textsc{Set}. 
In total, $VGPI$ consists of 825 images, 460 human instructions for pick, 68 human instructions for place, 116,999 self-generated instructions for pick, and 135,421 self-generated instructions for place. 

\noindent\textbf{Robotic system setup.} Our robotic platform consists of Kinova Gen3 lite for manipulation, Intel Realsense Depth Camera D435, and Azure Kinect DK for RGB images. The platform communicates with the remote workspace conducting VG and computes the manipulation planning locally.

\subsection{Offline Experiment}
\label{sec:offline}

\noindent\textbf{Evaluation protocol.} We evaluate \textsc{Get} with the Intersection over Union (\textsc{IoU}) score, which is a dominant metric in visual grounding (VG). The IoU compares the ground truth and the predicted bounding boxes and divides the overlapping regions between the two by their union. We report $precision@0.5$ scores, which measure the percentage of predicted regions with an IoU score greater than 0.5.

\noindent\textbf{Compared methods.} 
We report results in a combination of two different dimensions: (1) backbone VG models (i.e., OFA~\cite{ofa} and MDETR~\cite{mdetr}) and (2) three methods (i.e., Zero-Shot (Z-S in Tab.~\ref{tab:offline_result}), PseudoQ, and GVCCI). 
The Zero-Shot method (Z-S in Tab.~\ref{tab:offline_result}) is to evaluate backbone VG models (i.e., OFA~\cite{ofa} and MDETR~\cite{mdetr}) on the real-world environment without any adaptation.
The PseudoQ approach is to evaluate \textsc{Get} adapting to the real-world environment, using the instructions generated by PseudoQ.
To ensure a fair comparison, all modules and settings, except for instruction generation, were kept identical to those used in our framework.
All of the methods are evaluated on Test-H, Test-R, and Test-E datasets, and with four different time steps in log scale $\tau \in \{8,33,135,540\}$\footnote{At each chosen time step, the model is updated for an epoch with received images on the top of the former model (e.g., at time step 33, the model is updated on the top of the model from time step 8).} except for the Zero-Shot method\footnote{Time step for the Zero-Shot method is 0.}. 

\noindent\textbf{Results.} 
We present our results for the offline setting in Tab.~\ref{tab:offline_result}.
The results indicate that GVCCI outperforms the baseline methods in locating target objects across all test sets (Test-H, Test-R, and Test-E) and backbone models (OFA and MDETR).
It is worth noticing that GVCCI with $\tau=8$ in \textsc{Env1} (i.e., Test-H and Test-R) yields dramatic performance gains with only learning eight scenes compared with the Zero-Shot method. 
The performance monotonically increases as the observed scenes accumulate in the buffer. 
On the other hand, the PseudoQ method fails to demonstrate incremental improvements, indicating an early overfitting of the model.
It indicates that the instructions generated by PseudoQ make the VG model overfit with a small number of data. 
In \textsc{Env2} (i.e., Test-E), we observed relatively modest improvements in performance and even a performance degradation at $\tau=8$.
We noticed that Faster RCNN~\cite{fasterRCNN} performs poorly in the presence of noisy backgrounds in \textsc{Env2}, occasionally generating obstructive pseudo instructions.
However, when compared to PseudoQ, GVCCI exhibits superior performance, which can be attributed to its robustness to errors from Faster RCNN while generating higher-quality pseudo instructions.


\begin{figure*}[t]
\centering
\includegraphics[width=\linewidth]{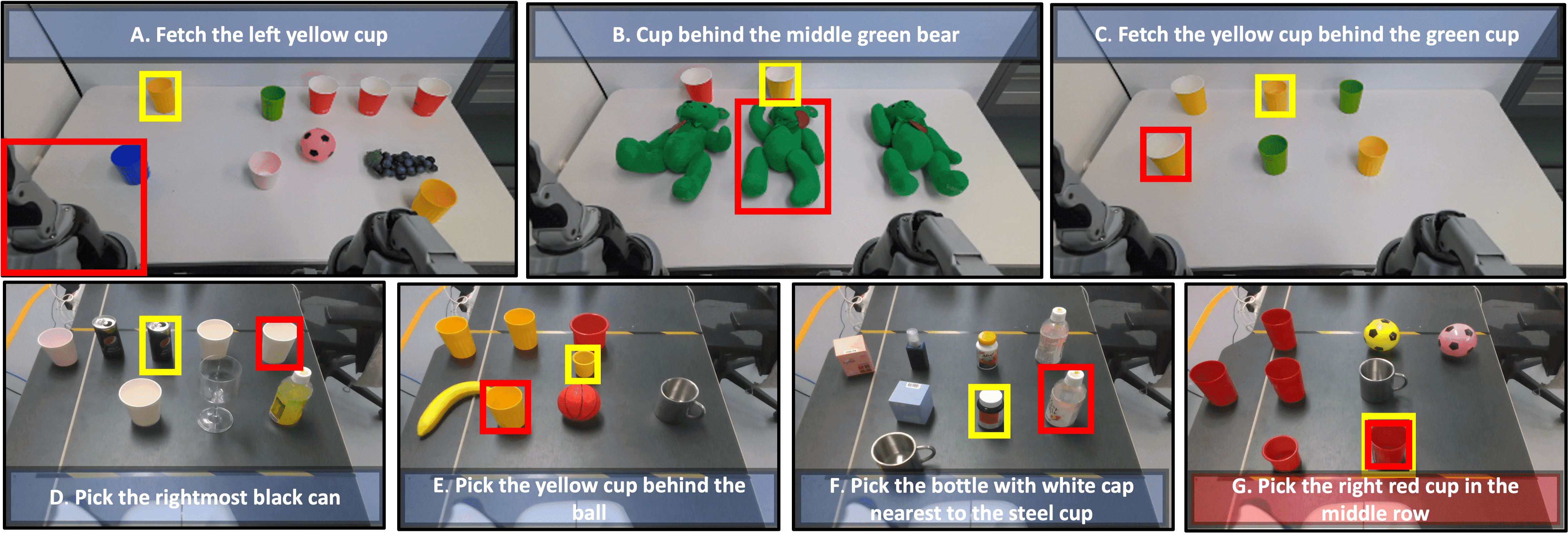}
\caption{\textbf{Examples of visual grounding results from \textsc{Env1} (above row) and from \textsc{Env2} (bottom row)}. Results of \textsc{Get} ($\tau=540$ on \textsc{Env1} and $\tau=135$ on \textsc{Env2}) and Zero-Shot results of OFA ($\tau=0$) are depicted in yellow and red bounding boxes, respectively.
Sentences are the pick instructions given where instructions in blue boxes gave a correct inference from \textsc{Get} and the instruction in the red box gave an incorrect inference.
Example images were extracted from our Test-R and Test-E datasets.
Note that instructions B, F, and G do not follow the templates that GVCCI uses.}
\label{fig:qualitative}
\end{figure*}

\subsection{Qualitative Analysis}
\label{ssec:qualitative}

We visualize examples of VG results from both \textsc{Get} and OFA~\cite{ofa} in Fig.~\ref{fig:qualitative} to analyze the cause of the considerable performance gain in Sec.~\ref{sec:offline}. 
We categorize the cause into three main reasons. 
First, we observed that \textsc{Get} performs high-dimensional inference. For example C, the model has to deduce that the green cup in the instruction means the left one.
Exploring these examples, we observe out-of-the-box reasoning capabilities from \textsc{Get} trained on self-generated instructions.
Second, we observe that OFA results in undesirable grounding due to some biases induced by the VG dataset
~\cite{refcoco}. For example, A and D in Fig.~\ref{fig:qualitative} show the bias of relational terms in OFA. Specifically, example D shows the extreme case of the bias, finding an utterly irrelevant object fooled by the term “rightmost.” 
Furthermore, as seen in example B, OFA seems to ground phrases that occupy a longer portion in the instruction, i.e., ground ``the middle green bear'' rather than ``Cup.'' 
Meanwhile, our method seems robust from these biases by learning with plentifully generated relational expressions from diverse object combinations. 
Finally, we discovered the effect of the domain discrepancy between RefCOCO and our environment on the model's prediction. 
Since the objects queried in RefCOCO have larger portions in the image than in our environment, OFA tends to ground objects that have a larger portion in the image, as seen in examples A, B, C, E, F, and G. In other words, OFA tends to choose a most probable object that ‘are to be’ queried. 
E shows that OFA neglects all the information except for the “yellow cup”, choosing the largest object. On the other hand, our method seems to relieve this problem by adapting to all the detected objects from our environment. 
In addition, A shows the bottleneck of inference in a robotic manipulation environment. 
Since OFA tends to ground larger objects, the model occasionally ground instructions to a robot arm that appeared in the robot's visual perception, which is inevitable in robotic manipulation tasks. 
Adaptation with its own perception helps a robot to learn that a part of its body is “not” a desired object to be chosen. 

\begin{figure}[t]
\centering
\includegraphics[width=\linewidth]{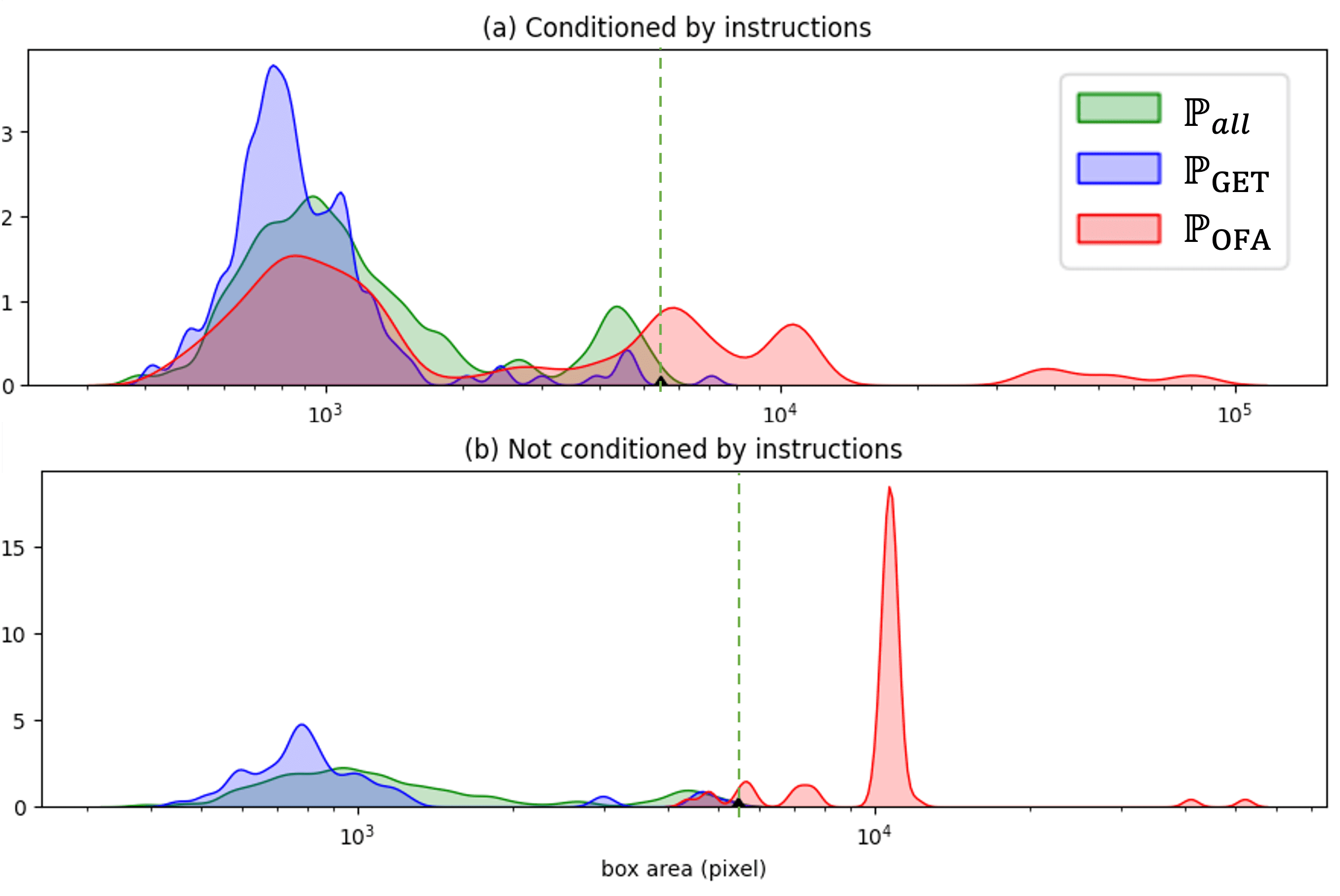}
\caption{\textbf{Empirical probability density functions of the size of objects.} Each row corresponds to a case of whether language instructions condition the predictions or not. $\mathbb{P}_{all}$ approximates the density of all object's size for all images in Test-H data, $\mathbb{P}_{\textsc{Get}}$ and $\mathbb{P}_{\textsc{OFA}}$ are each approximated densities of the predicted object areas predicted by \textsc{Get} at the time step 540 and by OFA. Dotted lines denote the maximal size of the object area in $\mathbb{P}_{all}$.}
\label{fig:o-size}
\end{figure}

To delve into the OFA's tendency of choosing a larger object, we visualize three probability density functions\footnote{Probability densities are approximated with Gaussian Kernel Density Estimation with smoothness factor of 3.} in Fig.~\ref{fig:o-size}: $\mathbb{P}_{\textsc{Get}}$ and $\mathbb{P}_{\textsc{OFA}}$, regarding the size of the object bounding box each predicted by \textsc{Get} and OFA, and $\mathbb{P}_{all}$, regarding the size of all the objects.
In addition, we visualize the density of objects that are predicted without instructional conditions to clearly observe the bias exhibited by OFA in Fig.~\ref{fig:o-size}-(b).
It is worth noting that the values of $\mathbb{P}_{\textsc{OFA}}$ extend beyond the maximum object size, indicating that when not conditioned by instructions, OFA is prone to infer large boxes that do not correspond to any objects. 
\textsc{Get}, on the other hand, seems to choose relatively diverse objects in the image regardless of size. 
In Fig.~\ref{fig:o-size}-(a), despite being conditioned by the instruction, OFA still tends to ground larger objects and even fails to box out the area with no objects in some cases.
We also compare the similarity of distributions with scaled Wasserstein distance, $W$. For (a), $W(\mathbb{P}_{\textsc{OFA}}, \mathbb{P}_{all})=47.73$ and $W(\mathbb{P}_{\textsc{Get}}, \mathbb{P}_{all})=4.69$. For (b), $W(\mathbb{P}_{\textsc{OFA}}, \mathbb{P}_{all})=95.21$ and $W(\mathbb{P}_{\textsc{Get}}, \mathbb{P}_{all})=3.44$. 
The distance $W(\mathbb{P}_{\textsc{OFA}}, \mathbb{P}_{all})$ and $W(\mathbb{P}_{\textsc{Get}}, \mathbb{P}_{all})$ in both (a) and (b) numerically tell that $\mathbb{P}_{\textsc{Get}}$ is more similar to $\mathbb{P}_{all}$ than $\mathbb{P}_{\textsc{OFA}}$ is, partially proving that \textsc{Get} is well adapted to the environment.

\subsection{Online Experiment}
We conduct Language-Guided Robotic Manipulation (LGRM) with a real-world arm robot loaded with \textsc{Get} and \textsc{Set}. The task is divided into four consecutive procedures: (1) inferring the accurate object location referred to by the instruction, (2) grasping the target object with its bounding box coordinates predicted by \textsc{Get}, (3) reasoning target location from the place instruction with \textsc{Set}, and (4) put the target object on the target location. We perform the online experiment by reconstructing the scenes in \textsc{Env2} test data. 

\noindent\textbf{Evaluation protocol.}
LGRM was evaluated on four checkpoints: a success rate of pick inference; pick manipulation; place inference; and overall pick-and-place accuracy. Evaluating pick manipulation is straightforward; we regard a correct one if the robot picks up the target object referred in the instruction. However, assessing the accuracy of pick-and-place operations can be subjective for several reasons. For instance, objects like a cup may remain lying down in the target location, leading to ambiguity regarding the decision to succeed. Additionally, people may have a different interpretation of the correct location for placing the object according to the given instruction. Therefore, we assess the accuracy of pick-and-place via human evaluation. Specifically, three participants made decisions based on (1) pick-and-place instruction and (2) images before and after the manipulation was held. Finally, the participants evaluated pick inference and place inference scores since the success criteria can differ from person to person. We report the average scores from all the participants.

\noindent\textbf{Compared methods.} Similar to the offline experiment, we compare GVCCI with two methods: (1) Zero-Shot and (2) PseudoQ~\cite{pseudoq}. However, as discussed in \ref{ssec:vgm}, the Zero-Shot method can not infer the target location to place since the target location is a vacant region in most cases. Thus, we built a simple rule-based framework for place operation. The framework first splits the instruction into three parts: command term (e.g., ``place it"); relation term (e.g., ``in behind"); and referring expression term (e.g., ``the yellow cup")\footnote{Annotators were guided to use relation terms among \{in front of, in behind, on the left/right side of\} when annotating place instructions for the valid comparison, making rule-based parser work at inference.}. The VG model then infers the bounding box of the reference object using the referring expression term and shifts the bounding box coordinates to a fixed extent considering the relation term. Consequently, we compare GVCCI with two methods: (1) a Zero-Shot method using OFA for pick and OFA with the rule-based method for place and (2) \textsc{Get} and \textsc{Set} trained with synthetic instructions from PseudoQ. Additionally, we benchmarked a variant of GVCCI, \textsc{Get} for pick and \textsc{Get} with a rule-based method for place operation.   


\begin{table}[t]
\caption{
\begin{flushleft}
\textnormal{
\textbf{Results of online experiments in \textsc{Env2}.}
Two VG models are used for each pick and place. Adaptation approaches are denoted in brackets: Zero-Shot (i.e., not adapted), PseudoQ, and GVCCI (ours). Place models with $\dagger$ denote the model using the rule-based method for place operation. We report the absolute performance gains compared with the Zero-Shot performance of OFA.
}
\end{flushleft}
}
\label{tab:online_result}
\begin{center}
\begin{tabular}{ccccc}
\toprule
\begin{tabular}[c]{@{}c@{}}\textbf{Pick/Place}\\ \textbf{Model}\end{tabular} 
& \begin{tabular}[c]{@{}c@{}}\textbf{Pick}\\ \textbf{Inference}\end{tabular} 
& \begin{tabular}[c]{@{}c@{}}\textbf{Pick}\\ \textbf{Manip.}\end{tabular} 
& \begin{tabular}[c]{@{}c@{}}\textbf{Place}\\ \textbf{Inference}\end{tabular} 
& \begin{tabular}[c]{@{}c@{}}\textbf{Pick}\\ \textbf{\&}\\ \textbf{Place}\end{tabular} 
\\
\midrule
 \begin{tabular}[c]{@{}c@{}}\textbf{\textsc{OFA}/\textsc{OFA{$^\dagger$}}}\\ \textbf{(zero-shot)}\end{tabular}
& \begin{tabular}[c]{@{}c@{}}51.47\\ \base\end{tabular}
& \begin{tabular}[c]{@{}c@{}}35.29\\ \base\end{tabular}
& \begin{tabular}[c]{@{}c@{}}69.12\\ \base\end{tabular}
& \begin{tabular}[c]{@{}c@{}}21.57\\ \base\end{tabular}
\\
\midrule
\begin{tabular}[c]{@{}c@{}} \textbf{\textsc{Get}/\textsc{Set}} \\ \textbf{(PseudoQ)}\end{tabular}
& \begin{tabular}[c]{@{}c@{}}64.71\\ \gap{13.24}\end{tabular}
& \begin{tabular}[c]{@{}c@{}}58.82\\ \gap{23.53}\end{tabular}
& \begin{tabular}[c]{@{}c@{}}73.03\\ \gap{3.91}\end{tabular}
& \begin{tabular}[c]{@{}c@{}}33.32\\ \gap{11.75}\end{tabular}
\\
\midrule
 \begin{tabular}[c]{@{}c@{}}\textbf{\textsc{Get}/\textsc{Get}$^\dagger$}\\ \textbf{(GVCCI)}\end{tabular}
& \begin{tabular}[c]{@{}c@{}}\textbf{82.35}\\ \gapbold{30.88}\end{tabular}
& \begin{tabular}[c]{@{}c@{}}72.06\\ \gap{36.77}\end{tabular}
& \begin{tabular}[c]{@{}c@{}}\textbf{77.94}\\ \gapbold{8.82}\end{tabular}
& \begin{tabular}[c]{@{}c@{}}47.06\\ \gap{25.49}\end{tabular}
\\
\midrule
\begin{tabular}[C]{@{}C@{}}
\textbf{\textsc{Get}/\textsc{Set}} \\ \textbf{(GVCCI)}
\end{tabular}
& \begin{tabular}[C]{@{}C@{}}80.88\\ \gap{29.41}\end{tabular}
& \begin{tabular}[C]{@{}C@{}}\textbf{75.00}\\ \gapbold{39.71}\end{tabular}
& \begin{tabular}[C]{@{}C@{}}76.47\\ \gap{7.35}\end{tabular}
& \begin{tabular}[C]{@{}C@{}}\textbf{50.98}\\ \gapbold{29.41}\end{tabular}
\\
\bottomrule
\end{tabular}
\end{center}
\end{table}

\noindent\textbf{Results.} 
As shown in Tab.~\ref{tab:online_result}, \textsc{Get} and \textsc{Set} trained with GVCCI remarkably outperform both OFA and \textsc{Get} and \textsc{Set} trained with PseudoQ~\cite{pseudoq} on all checkpoints.
Mainly, our method showed 29.41\% absolute points gain of the overall Pick\&Place accuracy as to the OFA.
The robot using OFA only achieved 35.29\% success rate of pick manipulation (-16.18\% absolute points as to the Pick Inference accuracy), often failing to grasp the correct object that the model inferred successfully.
On the other hand, the robot using \textsc{Get} achieved 73.53\% success rate of pick manipulation on average (-8.13\% absolute points as to the average Pick Inference accuracy), relatively grasping the objects better from \textsc{Get} than from OFA.
We analyzed that the main reason is from the tighter bounding box that OFA yielded, as shown in Fig.~\ref{fig:qualitative}-G. 
Since leveraging all the information of the object point cloud plays a dominant role in manipulating the object, loss of the information due to the tighter bounding box can lead to failure in manipulation. 
On the other hand, \textsc{Get} inferred a looser bounding box, fully exploiting object point cloud information leading to less failure in picking manipulation.
Inferring the target destination point from place instructions was best using \textsc{Get} with a rule-based method. However, \textsc{Set} gave a comparable performance, nearly achieving \textsc{Get} with the rule-based method. 
Since the rule-based method can only handle the place instructions that use the exact relation term provided, \textsc{Set} is expected to give better results when handling free-form place instructions.
To summarize the results, our proposed approach enhances the capability to comprehend human instructions and perform manipulations with greater precision, compared to both trained methods (i.e., PseudoQ) and non-trained methods (i.e., Zero-Shot OFA).

\section{CONCLUSION}

This paper presents a novel lifelong learning framework for visual grounding in object-centric robotic manipulation tasks without human supervision.
Our framework significantly enhances visual grounding in three real-world pick-and-place scenarios, surpassing the performance of state-of-the-art visual grounding methods and enabling robots to follow human instructions more accurately. 
Although our work does not claim to capture all possible human instructions, we believe it represents a valuable step towards making intelligent robots capable of understanding natural language instructions and becoming omnipresent in our daily lives.




\section*{Acknowledgements}
We extend our gratitude to Jungmin Lee for her invaluable contribution to video editing, Suhyung Choi for assistance in data collection, and Minjoon Jung for valuable suggestions on the writing. We also express our appreciation to all the reviewers for their insightful comments and feedback.
\bibliographystyle{IEEEtran.bst}
\bibliography{ref}

\begin{thebibliography}{10}
\providecommand{\url}[1]{#1}
\csname url@samestyle\endcsname
\providecommand{\newblock}{\relax}
\providecommand{\bibinfo}[2]{#2}
\providecommand{\BIBentrySTDinterwordspacing}{\spaceskip=0pt\relax}
\providecommand{\BIBentryALTinterwordstretchfactor}{4}
\providecommand{\BIBentryALTinterwordspacing}{\spaceskip=\fontdimen2\font plus
\BIBentryALTinterwordstretchfactor\fontdimen3\font minus
  \fontdimen4\font\relax}
\providecommand{\BIBforeignlanguage}[2]{{%
\expandafter\ifx\csname l@#1\endcsname\relax
\typeout{** WARNING: IEEEtran.bst: No hyphenation pattern has been}%
\typeout{** loaded for the language `#1'. Using the pattern for}%
\typeout{** the default language instead.}%
\else
\language=\csname l@#1\endcsname
\fi
#2}}
\providecommand{\BIBdecl}{\relax}
\BIBdecl

\bibitem{refcoco}
L.~Yu, P.~Poirson, S.~Yang, A.~C. Berg, and T.~L. Berg, ``Modeling context in
  referring expressions,'' in \emph{European Conference on Computer
  Vision}.\hskip 1em plus 0.5em minus 0.4em\relax Springer, 2016, pp. 69--85.

\bibitem{referitgame}
S.~Kazemzadeh, V.~Ordonez, M.~Matten, and T.~Berg, ``Referitgame: Referring to
  objects in photographs of natural scenes,'' in \emph{Proceedings of the 2014
  conference on empirical methods in natural language processing (EMNLP)},
  2014, pp. 787--798.

\bibitem{ofa}
P.~Wang, A.~Yang, R.~Men, J.~Lin, S.~Bai, Z.~Li, J.~Ma, C.~Zhou, J.~Zhou, and
  H.~Yang, ``Ofa: Unifying architectures, tasks, and modalities through a
  simple sequence-to-sequence learning framework,'' in \emph{International
  Conference on Machine Learning}.\hskip 1em plus 0.5em minus 0.4em\relax PMLR,
  2022, pp. 23\,318--23\,340.

\bibitem{mdetr}
A.~Kamath, M.~Singh, Y.~LeCun, G.~Synnaeve, I.~Misra, and N.~Carion,
  ``Mdetr-modulated detection for end-to-end multi-modal understanding,'' in
  \emph{Proceedings of the IEEE/CVF International Conference on Computer
  Vision}, 2021, pp. 1780--1790.

\bibitem{vilbert}
J.~Lu, D.~Batra, D.~Parikh, and S.~Lee, ``Vilbert: Pretraining task-agnostic
  visiolinguistic representations for vision-and-language tasks,''
  \emph{Advances in neural information processing systems}, vol.~32, 2019.

\bibitem{refcocog}
J.~Mao, J.~Huang, A.~Toshev, O.~Camburu, A.~L. Yuille, and K.~Murphy,
  ``Generation and comprehension of unambiguous object descriptions,'' in
  \emph{Proceedings of the IEEE conference on computer vision and pattern
  recognition}, 2016, pp. 11--20.

\bibitem{srfnlifrm}
S.~G. Venkatesh, A.~Biswas, R.~Upadrashta, V.~Srinivasan, P.~Talukdar, and
  B.~Amrutur, ``Spatial reasoning from natural language instructions for robot
  manipulation,'' in \emph{2021 IEEE International Conference on Robotics and
  Automation (ICRA)}.\hskip 1em plus 0.5em minus 0.4em\relax IEEE, 2021, pp.
  11\,196--11\,202.

\bibitem{invigorate}
H.~Zhang, Y.~Lu, C.~Yu, D.~Hsu, X.~La, and N.~Zheng, ``Invigorate: Interactive
  visual grounding and grasping in clutter,'' \emph{arXiv preprint
  arXiv:2108.11092}, 2021.

\bibitem{ingress}
M.~Shridhar, D.~Mittal, and D.~Hsu, ``Ingress: Interactive visual grounding of
  referring expressions,'' \emph{The International Journal of Robotics
  Research}, vol.~39, no. 2-3, pp. 217--232, 2020.

\bibitem{lopfribhsr}
O.~Mees, A.~Emek, J.~Vertens, and W.~Burgard, ``Learning object placements for
  relational instructions by hallucinating scene representations,'' in
  \emph{2020 IEEE International Conference on Robotics and Automation
  (ICRA)}.\hskip 1em plus 0.5em minus 0.4em\relax IEEE, 2020, pp. 94--100.

\bibitem{cpaptbgl}
O.~Mees and W.~Burgard, ``Composing pick-and-place tasks by grounding
  language,'' in \emph{Experimental Robotics: The 17th International
  Symposium}.\hskip 1em plus 0.5em minus 0.4em\relax Springer, 2021, pp.
  491--501.

\bibitem{avgrefrm}
Y.~Wang, K.~Wang, Y.~Wang, D.~Guo, H.~Liu, and F.~Sun, ``Audio-visual grounding
  referring expression for robotic manipulation,'' in \emph{2022 International
  Conference on Robotics and Automation (ICRA)}.\hskip 1em plus 0.5em minus
  0.4em\relax IEEE, 2022, pp. 9258--9264.

\bibitem{ivgorefhri}
M.~Shridhar and D.~Hsu, ``Interactive visual grounding of referring expressions
  for human-robot interaction,'' \emph{arXiv preprint arXiv:1806.03831}, 2018.

\bibitem{gssrefhri}
------, ``Grounding spatio-semantic referring expressions for human-robot
  interaction,'' \emph{arXiv preprint arXiv:1707.05720}, 2017.

\bibitem{reiofitsf}
D.~Whitney, E.~Rosen, J.~MacGlashan, L.~L. Wong, and S.~Tellex, ``Reducing
  errors in object-fetching interactions through social feedback,'' in
  \emph{2017 IEEE International Conference on Robotics and Automation
  (ICRA)}.\hskip 1em plus 0.5em minus 0.4em\relax IEEE, 2017, pp. 1006--1013.

\bibitem{plummer2015flickr30k}
B.~A. Plummer, L.~Wang, C.~M. Cervantes, J.~C. Caicedo, J.~Hockenmaier, and
  S.~Lazebnik, ``Flickr30k entities: Collecting region-to-phrase
  correspondences for richer image-to-sentence models,'' in \emph{Proceedings
  of the IEEE international conference on computer vision}, 2015, pp.
  2641--2649.

\bibitem{iprwowusli}
J.~Hatori, Y.~Kikuchi, S.~Kobayashi, K.~Takahashi, Y.~Tsuboi, Y.~Unno, W.~Ko,
  and J.~Tan, ``Interactively picking real-world objects with unconstrained
  spoken language instructions,'' in \emph{2018 IEEE International Conference
  on Robotics and Automation (ICRA)}.\hskip 1em plus 0.5em minus 0.4em\relax
  IEEE, 2018, pp. 3774--3781.

\bibitem{fasterRCNN}
S.~Ren, K.~He, R.~Girshick, and J.~Sun, ``Faster r-cnn: Towards real-time
  object detection with region proposal networks,'' \emph{Advances in neural
  information processing systems}, vol.~28, 2015.

\bibitem{bottomup}
P.~Anderson, X.~He, C.~Buehler, D.~Teney, M.~Johnson, S.~Gould, and L.~Zhang,
  ``Bottom-up and top-down attention for image captioning and visual question
  answering,'' in \emph{Proceedings of the IEEE conference on computer vision
  and pattern recognition}, 2018, pp. 6077--6086.

\bibitem{mattnet}
L.~Yu, Z.~Lin, X.~Shen, J.~Yang, X.~Lu, M.~Bansal, and T.~L. Berg, ``Mattnet:
  Modular attention network for referring expression comprehension,'' in
  \emph{Proceedings of the IEEE Conference on Computer Vision and Pattern
  Recognition}, 2018, pp. 1307--1315.

\bibitem{transformer}
A.~Vaswani, N.~Shazeer, N.~Parmar, J.~Uszkoreit, L.~Jones, A.~N. Gomez,
  {\L}.~Kaiser, and I.~Polosukhin, ``Attention is all you need,''
  \emph{Advances in neural information processing systems}, vol.~30, 2017.

\bibitem{pseudoq}
H.~Jiang, Y.~Lin, D.~Han, S.~Song, and G.~Huang, ``Pseudo-q: Generating pseudo
  language queries for visual grounding,'' in \emph{Proceedings of the IEEE
  Conference on Computer Vision and Pattern Recognition}, 2022.

\bibitem{ocidref}
K.-J. Wang, Y.-H. Liu, H.-T. Su, J.-W. Wang, Y.-S. Wang, W.~H. Hsu, and W.-C.
  Chen, ``Ocid-ref: A 3d robotic dataset with embodied language for clutter
  scene grounding,'' \emph{arXiv preprint arXiv:2103.07679}, 2021.

\bibitem{saycan}
M.~Ahn, A.~Brohan, N.~Brown, Y.~Chebotar, O.~Cortes, B.~David, C.~Finn,
  K.~Gopalakrishnan, K.~Hausman, A.~Herzog \emph{et~al.}, ``Do as i can, not as
  i say: Grounding language in robotic affordances,'' \emph{arXiv preprint
  arXiv:2204.01691}, 2022.

\bibitem{aabfigifet}
M.~Shridhar, J.~Thomason, D.~Gordon, Y.~Bisk, W.~Han, R.~Mottaghi,
  L.~Zettlemoyer, and D.~Fox, ``Alfred: A benchmark for interpreting grounded
  instructions for everyday tasks,'' in \emph{Proceedings of the IEEE/CVF
  conference on computer vision and pattern recognition}, 2020, pp.
  10\,740--10\,749.

\bibitem{gsrfhri}
S.~Guadarrama, L.~Riano, D.~Golland, D.~Go, Y.~Jia, D.~Klein, P.~Abbeel,
  T.~Darrell \emph{et~al.}, ``Grounding spatial relations for human-robot
  interaction,'' in \emph{2013 IEEE/RSJ International Conference on Intelligent
  Robots and Systems}.\hskip 1em plus 0.5em minus 0.4em\relax IEEE, 2013, pp.
  1640--1647.

\bibitem{nlcwr}
Y.~Bisk, D.~Yuret, and D.~Marcu, ``Natural language communication with
  robots,'' in \emph{Proceedings of the 2016 Conference of the North American
  Chapter of the Association for Computational Linguistics: Human Language
  Technologies}, 2016, pp. 751--761.

\bibitem{cwrumrn}
B.~Pi{\v{s}}l and D.~Mare{\v{c}}ek, ``Communication with robots using
  multilayer recurrent networks,'' in \emph{Proceedings of the First Workshop
  on Language Grounding for Robotics}, 2017, pp. 44--48.

\bibitem{iogusg}
J.~S.~K. Yi, Y.~Kim, and S.~Chernova, ``Incremental object grounding using
  scene graphs,'' \emph{arXiv preprint arXiv:2201.01901}, 2022.

\bibitem{sim2real}
G.~Tziafas and H.~Kasaei, ``Sim-to-real transfer of visual grounding for
  human-aided ambiguity resolution,'' \emph{arXiv preprint arXiv:2205.12089},
  2022.

\bibitem{visual_genome}
R.~Krishna, Y.~Zhu, O.~Groth, J.~Johnson, K.~Hata, J.~Kravitz, S.~Chen,
  Y.~Kalantidis, L.-J. Li, D.~A. Shamma \emph{et~al.}, ``Visual genome:
  Connecting language and vision using crowdsourced dense image annotations,''
  \emph{International journal of computer vision}, vol. 123, pp. 32--73, 2017.

\bibitem{wahl2005identifying}
R.~Wahl, M.~Guthe, and R.~Klein, ``Identifying planes in point-clouds for
  efficient hybrid rendering,'' in \emph{The 13th Pacific Conference on
  Computer Graphics and Applications}, vol.~3, 2005.

\bibitem{coleman2014moveit}
D.~Coleman, I.~Sucan, S.~Chitta, and N.~Correll, ``Reducing the barrier to
  entry of complex robotic software: a moveit! case study,'' \emph{arXiv
  preprint arXiv:1404.3785}, 2014.

\end{thebibliography}

\end{document}